\documentclass{article}

\usepackage[letterpaper,top=30truemm,bottom=30truemm,left=25truemm,right=25truemm]{geometry}
\usepackage[numbers]{natbib}
\usepackage[utf8]{inputenc}
\usepackage[T1]{fontenc}   
\usepackage{fourier}
\usepackage{hyperref}       
\usepackage{url}           
\usepackage{booktabs}       
\usepackage{amsfonts}       
\usepackage{nicefrac}       
\usepackage{microtype}      
\usepackage{placeins}

\usepackage[page]{appendix}
\usepackage{bm}
\usepackage{xspace}
\RequirePackage{algorithm}
\RequirePackage{algpseudocode}
\usepackage{subcaption}
\usepackage{wrapfig}
\usepackage{subcaption} 
\usepackage{multirow}
\usepackage[export]{adjustbox}
\usepackage{amsmath}
\usepackage{graphicx}
\usepackage{xcolor}
\usepackage{tikz}
\usetikzlibrary{calc} 
\usepackage{standalone}
\usepackage[version=4]{mhchem}
\usepackage{tabularx}
\usepackage{doi}
\usepackage[textsize=small, textwidth=35mm]{todonotes}

\newcommand{\figDate}{20240717-1240}  

\DeclareCaptionSubType[Alph]{figure}

\newcommand{\fig}[1]{Fig.~\ref{#1}}  
\newcommand{\nr}[1]{``\emph{#1}''}   % for namerxn classes

\newcommand{\trainingCode}{~\url{https://github.com/john-bradshaw/rxn-lm}}  
\newcommand{\splitCode}{~\url{https://github.com/john-bradshaw/rxn-splits}}  

\definecolor{act1}{HTML}{784479}
\definecolor{act2}{HTML}{1E5B29}
\definecolor{grayRandom}{HTML}{8b959e} 
\definecolor{blueOOD}{HTML}{002896}  
\definecolor{idGreen}{HTML}{43893D}
\definecolor{oodRed}{HTML}{D24C2B}

\newcommand{\affil}[1]{\ifcase#1\or$\dagger$\or$\sharp$\or$\ddagger$\or$\mathsection$\or$\clubsuit$\or$\mathparagraph$\fi}
\newcommand{\saffil}[1]{\textsuperscript{\affil{#1}}}

\title{Challenging reaction prediction models \\to generalize to novel chemistry}

\author{
John Bradshaw\saffil{1}
\hspace{0.5cm} 
Anji Zhang\saffil{1}
\hspace{0.5cm}
Babak Mahjour\saffil{1}
\hspace{0.5cm}
David E. Graff\saffil{2}\saffil{1} \\
\hspace{0.5cm}
Marwin H.S. Segler\saffil{4}
\hspace{0.5cm}
Connor W. Coley\saffil{1}\saffil{3}  \\
\small \affil{1} Department of Chemical Engineering, Massachusetts Institute of Technology (\texttt{\{jbrad, anji\_z, bmahjour, ccoley\}@mit.edu}); \\
 \small \affil{2} Department of Chemistry and Chemical Biology, Harvard University (\texttt{deg711@g.harvard.edu}); \\
  \small \affil{4} Microsoft Research AI for Science; \\
 \small \affil{3} Department of Electrical Engineering and Computer Science, Massachusetts Institute of Technology.
}

\date{}

\begin{document}

\maketitle

\begin{abstract}%

\noindent Deep learning models for anticipating the products of organic reactions have found many use
cases, including validating retrosynthetic pathways and constraining synthesis-based molecular design tools. 
Despite compelling performance on popular benchmark tasks, strange and erroneous predictions sometimes ensue when using these models in practice.
The core issue is that common 
benchmarks test models in an \emph{in-distribution} setting, whereas many real-world uses for these models are
in \emph{out-of-distribution} settings and require a greater degree of extrapolation. 
To better understand how current reaction predictors work in out-of-distribution domains, we report a
series of more challenging evaluations of a prototypical SMILES-based deep learning model.  
First, we illustrate how performance on randomly sampled datasets is overly optimistic compared to performance when generalizing to new patents or new authors. 
Second, we conduct time splits that evaluate how models perform when tested on reactions published in years after those in their training set, mimicking real-world deployment. 
Finally, we consider extrapolation across reaction classes to reflect what would be required for the discovery of novel reaction types. 
This panel of tasks can reveal the capabilities and limitations of today's reaction predictors, acting as 
 a crucial first step in the
development of tomorrow's next-generation models capable of reaction discovery.
\end{abstract}

\section{Introduction}%
\begin{figure}[t]
\centering
\includegraphics[width=0.9\textwidth]{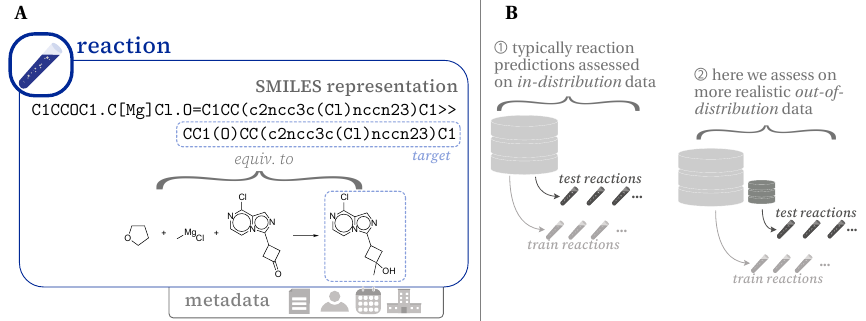}

\begin{subcaptiongroup}
    \phantomcaption\label{fig:overview:reactionPrediction}
    \phantomcaption\label{fig:overview:OOD}
\end{subcaptiongroup}
\captionsetup{subrefformat=parens}
\caption{\textbf{\subref{fig:overview:reactionPrediction}} 
Reaction prediction, in the context of this manuscript, is the task of predicting the major product(s) of a reaction given the reactants. 
(Note that by ``reaction'' we mean specific reported reaction examples, rather than generic reaction ``types'' or ``classes'' that cover a large group of related specific examples—we will come back to the concept of reaction types in Section~\ref{sect:nameRxn}.)
Chemical reaction datasets are often curated from academic or patent literature, and so each reaction is associated with a set of hidden metadata (the predictive model does not see this), such as the reaction's associated patent document, its authors, its publication date, the assignee/organization that filed the patent, etc. 
\textbf{\subref{fig:overview:OOD}} Typically, reaction predictors are assessed in an in-distribution setting, meaning that the training and test reactions come from the same distributions. 
However, in the real-world, reaction prediction models are often deployed on out-of-distribution data, a setup that we will discuss how to replicate.
 }
\end{figure}

Reaction prediction---the task of anticipating \textit{in silico} the products of a chemical reaction given the reactants (\fig{fig:overview:reactionPrediction}; \citep{Warr2014-sx,Goodman2018-op,Schwaller2019-ra,Tu2023-rl})---is a crucial technology in  (a)  the validation of retrosynthetic pathways \citep{ASKCOS-Team2019-fv,Schwaller2020-qg,Zhong2023-ko,Jaume-Santero2023-aj}, (b) as a component of synthesis-based de novo design algorithms \citep{Gao2022-vn,Bradshaw2020-bw,Chevillard2015-vt,Vinkers2003-ia,Korovina2020-bw,Gottipati2020-pi,Horwood2020-af,Swanson2024-gj,Bradshaw2019-ah,Wang2022-an}, and potentially (c) for the discovery of new reactions \citep{Bort2021-xm,Herges1990-jd,Segler2017-gs,Wang2022-ws,Mahjour2024-fx}.
Encouragingly, there has been a burst of recent works developing a variety of machine learning–based reaction predictors that achieve very high accuracies on common benchmark tasks \citep{Schwaller2019-ra,Coley2019-mu,Jin2017-hh,Sacha2021-lj,Bi2021-xm,Tu2022-xr,Bradshaw2019-cj,Wei2016-ro,Irwin2022-eq,Do2019-pn,Meng2023-be,Tetko2020-yb,Fooshee2018-fr,Kayala2012-dl}.
With the best of these models matching or outperforming human chemists (see, e.g., \citep[\S4.2]{Jin2017-hh}) and reporting top-5 accuracies above 95\% (meaning that the correct answer is found in the top five predictions of the model over 95\% of the time; see, e.g., \citep[p.9]{Tetko2020-yb}), performance seems to have saturated. Distinguishing best-performing models has become challenging. It is also natural to wonder if the task of reaction prediction has been ``solved'' to a meaningful degree.

When using these models in practice, it quickly becomes apparent that the answer is a resounding no. 
In fact, when using reaction predictors in new domains, not only might a model make an incorrect prediction, it might hallucinate a product preposterous to a human chemist. 
The discrepancy between the reported performance on benchmarks with the subjective performance that can be seen in practice can be explained by the setting in which the model is evaluated. 
Benchmark tasks (such as \texttt{USPTO\_Stereo}, \texttt{USPTO\_MIT}, Pistachio, etc. \citep{Lowe2012-ni, Schwaller2018-rh, Jin2017-hh, Mayfield2017-qy}) evaluate models on in-distribution (ID) data, where the reactions in the test set come from the same distribution as that used to train the model, for example, using a random partition of a reaction dataset. 
However, in practice we often want to evaluate a model on out-of-distribution (OOD) data, meaning the test reactions are sampled from a different distribution than that used to train the model (\fig{fig:overview:OOD}). In fact, using these models for reaction discovery is by definition an out-of-distribution task. 

The unrealistic nature of current evaluations not only robs us of a sense of how existing methods perform, but it does so in such a way that overstates performance, stymieing analysis of where methods fall short and how to improve them. 
To address this, we reassess what it means to evaluate a reaction predictor. We discuss and develop new tasks to test how well reaction predictors can do in different out-of-distribution domains, investigating when and how they are able to generalize and extrapolate in such settings. 
Concretely, we seek to answer the following questions:
\begin{enumerate}
    \item How over-optimistic are the random splits that are currently the most popular style of split for this task, and what is a more realistic evaluation of a reaction predictor's performance?
    \item If we want to use reaction predictors trained today on future datasets, how should we design benchmarks to test models prospectively?
    \item When, and under what circumstances, might reaction predictors be able to discover new reactions?
\end{enumerate}

\section{Results}

\subsection{Random splits are over-optimistic by ignoring dataset structure}
\label{sect:docAuth}

Typically, reaction predictors are tested on \emph{random splits}. 
That is, we treat a large reaction dataset (usually extracted from the patent literature \citep{Lowe2012-ni,Mayfield2017-qy,NextMove-Software2021-td}) as independently and identically distributed and randomly divide the reactions up between the training, validation, and test sets. 
The reactions in the training and validation sets are used for deciding on hyperparameters and for training the model (in what follows we will make no further distinction between the training and validation sets), while those in the test set are used for the final evaluation. 

\begin{figure}
    \centering
    \includegraphics[width=0.9\linewidth]{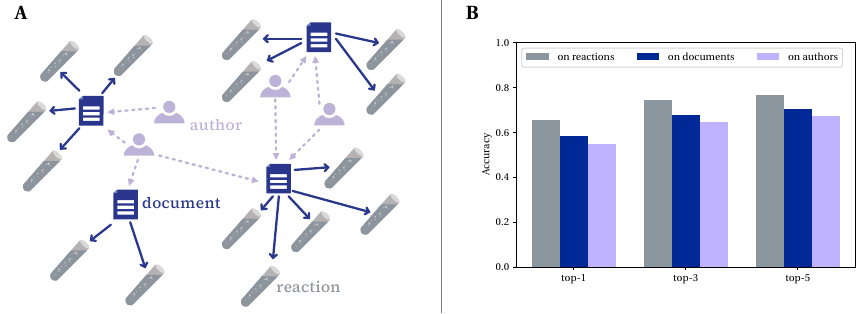}
\begin{subcaptiongroup}
    \phantomcaption\label{fig:docAuth:cartoon}
    \phantomcaption\label{fig:docAuth:results}
\end{subcaptiongroup}
\captionsetup{subrefformat=parens}
\caption{
    \textbf{\subref{fig:docAuth:cartoon}} Reaction datasets are formed by authors coming together and writing documents, which contain many  (often similar) reactions. 
    Evaluating a reaction predictor on train/test sets that account for this structure provide different accuracy scores.
    (Note that in this paper we clean and deduplicate reactions before creating the splits, such that a reaction is only associated with one document---see Section~\ref{sect:dataset}.)
    \textbf{\subref{fig:docAuth:results}} Top-1, 3, \& 5 accuracies when doing reaction- (i.e., random), document-, and author-based splits.
}
\label{fig:docAuth}
\end{figure}

However, the universe does not actually generate the reactions in reaction datasets independently!
Instead it generates chemists, who join organizations, form teams, and write documents (e.g., journal articles, patents, etc.) containing reactions (\fig{fig:docAuth:cartoon}). 
Often these documents contain many related reactions, for instance, a group of different reactants undergoing the same transform in an exploration of reaction scope or structure-activity relationships (SAR).
By creating a random split, many of these highly related reactions will be spread across the train and test sets.
This in turn means that
when making test predictions, the model can take advantage of the highly similar reactions that ended up in its training set, a scenario less likely to occur if one were truly independently sampling reactions for testing.

In order to better understand how this dataset structure affects the accuracy metrics of reaction predictors, we investigate alternative splits of the Pistachio dataset \citep{NextMove-Software2021-td,Mayfield2017-qy} that take this structure into account.
In particular, we compare three different strategies for dividing reactions into the training and test splits:
 (1) \emph{on reactions}, which is the same as a typical random splitting strategy; 
 (2) \emph{on documents}, which means that all reactions associated with each document end up together either in the training or test set;
 and (3) \emph{on authors}, which is similar to the document-based approach but done on authors instead, such that each author (and their corresponding reactions) is associated with either the training or test set.
 We train separate language-based reaction predictor models on each of these splits, controlling for dataset size, and present the accuracy results in \fig{fig:docAuth} (and Table~\ref{tab:docAuth}).
Specifically, we use an encoder-decoder Transformer model \citep{Vaswani2017-se}, based on the BART architecture \citep{Lewis2020-nf}. 
 In practice, this BART model is very similar to the Molecular Transformer model \citep{Schwaller2019-ra}, with a few small architectural differences. 
  Full details of the model and the training setup can be found in Section~\ref{sect:methods}. Evaluation focuses on the top-$k$ accuracy metric, which asks whether the experimentally-recorded major product appears in the $k$ highest ranked predictions by the model.

\fig{fig:docAuth} confirms that traditional, random splits (\emph{on reactions}) are over-optimistic relative to splitting on documents or authors. 
We see that a model trained and evaluated on an \emph{on reactions} split obtains a top-1 accuracy of $65$\%.
When instead splitting on documents, the same model obtains a lower accuracy of $58$\%, which further drops to $55$\% when splitting on author.
Similar trends are also found when looking at the top-3 and top-5 accuracies.
Overall, this indicates that the group of similar reactions associated with the same document or author leads to better reported model performance when these reactions are spread across both the training and test sets.
When using a reaction predictor ``in the wild,'' one is unlikely to be evaluating on reactions that are in a document already used to train the model, and so these document- and author-based splits are more likely to represent real-world performance. 
The drop of $\sim10$\% accuracy (on author-based splits) is therefore important not only in giving a better sense of current ML-based reaction predictor performance, but also in highlighting that there is still more room left for improvement than suggested by previous benchmarks.

\subsection{Time-based splits enable prospective evaluation mimicking real-world use}
\label{sect:timeSplits}
The document- and author-based splits considered in the previous section help provide a stricter evaluation of reaction predictors when they are used \emph{retrospectively}, i.e., to make predictions for inputs similar  
to already discovered and documented reactions. 
This can be the case, for instance, when checking the credibility of reactions in proposed synthesis plans, as the reactions will often be close (i.e., having similar reactants or similar transforms) to those already known, particularly if one wants confidence in their practicality.
However, often it's also important to know how well reaction predictors work \emph{prospectively}, i.e., on  reactions of interest going forward, which might involve a different distribution of reaction types or substrates. 
For example, when assessing the utility of reaction predictors for reaction discovery, one only wants to know how well they would work on new, undiscovered transforms.

\begin{figure}
\centering
\includegraphics[width=\linewidth]{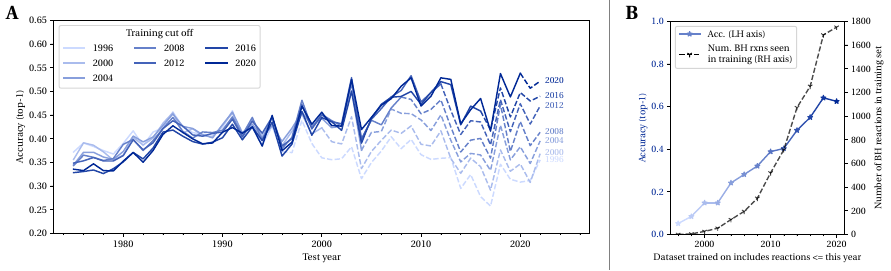}
\begin{subcaptiongroup}
    \phantomcaption\label{fig:timeSplits}
    \phantomcaption\label{fig:bhOverTime}
\end{subcaptiongroup}
\captionsetup{subrefformat=parens}
\caption{\textbf{\subref{fig:timeSplits}} Top-1 accuracies of reaction predictors trained up to different time cutoffs (different colors) when evaluated on held-out test sets for each year (x-axis). 
For instance, the line in the lightest shade, marked ``1996'', reports the top-1 accuracy for a reaction predictor trained on reactions that were reported up to 1996 (inclusive).
The dashed line indicates model performance when the model is ``extrapolating''---meaning that the test set year is beyond the model's time cutoff.
Note that we control for training set size so each model sees the same number of reactions in training (the absolute performance of the model is therefore lower than when training on all available data up to a given year). 
Further details on experimental setup and additional results can be found in Section~\ref{sect:timeBasedSplitsMethods} and \ref{sect:moreTimeSplits}.
\textbf{\subref{fig:bhOverTime}} Performance of the models trained on different time splits on a separate, static test set of Buchwald–Hartwig reactions. 
The blue solid line shows the top-1 accuracies (left-hand axis), while the dotted gray line shows the number of Buchwald–Hartwig reactions in the models'
training sets (right-hand axis).
 }
\label{fig:timeANDBHSplit}
\end{figure}

To evaluate how well reaction predictors work prospectively, we create a \emph{time-based} split.
Rather than just considering a single time-based split, as is typical \citep{Sheridan2013-ys}, we instead create a series of splits to  consider how model accuracy changes as the difference in time between the training and test set increases.
Specifically, this process works as follows:
 First of all, we take the same processed Pistachio dataset used in the previous section, which contains reactions from patents as early as the 1970s, and split off a separate held-out test set for each year.
Next, we take the remaining reaction data and create a sequence of training sets with different time cutoffs, each only containing reactions recorded up to and including in the associated cutoff year. 
Finally, we evaluate models trained on these different training sets on our held-out test sets (\fig{fig:timeSplits}). 
When creating these training sets, we control for training set size (results without this restriction are shown in Section~\ref{sect:moreTimeSplits}).

 \fig{fig:timeSplits} shows several important trends. 
Looking first at the performance of a single model in the interpolative regime, we see that while there is some inter-year variation---in part due to variability from the random selection of test sets---performance  gradually increases until the model reaches its training cutoff point.
 We hypothesize that this effect is due to the shifts in the distributions of reaction types reported over time---the reaction types present in later years are more popular (see \fig{fig:reactionCounts} or \citep{Schneider2016-lq,Roughley2011-pi}) and so better predicted.
 
 A second thing we notice in the interpolative regime is that the models trained on earlier cutoffs do better on the earlier years (particularly from 1975–1990).
This effect is linked with us controlling for the training set size: The models with the earlier cutoffs will have better specialized on reaction types present in the earlier time period because their fixed data budget is spread across a smaller range of years.
 When we remove this control (see \fig{fig:timeSplitsNoFix1} in Section~\ref{sect:moreTimeSplits}) and train each model on all of the data available up to its cutoff, this second trend disappears.
 
Switching to analyzing the extrapolative regime (indicated by the dashed line), we notice that  performance starts falling after the model's training cutoff.
Moreover, the drop in performance seems to be correlated with the difference between the training cutoff and test year, such that the larger the difference, the worse the performance.
As such, we can think of the time axis as acting as a proxy for an ``extrapolation distance.''
However, even though accuracy drops, the model trained on reactions reported up to 1996 still gets some predictions correct nearly 25 years out.
This is likely because there are plenty of reactions (such as the Suzuki coupling) that are popular and remain popular; overall, this suggests that current methods have some utility in predicting the products of reactions conducted in the future. 

\subsubsection{Reaction discovery and adoption in a time-based split}
\label{sect:buchwaldHartwigAnalysis}
One factor that the extrapolation distance represents in a time-based split is the discovery and then widespread adoption of a new reaction; this is part of the reason why extrapolating into the future may be hard. 
We can assess this further by looking at the performance of the models trained on the time-based splits on specific reactions discovered during the time horizon we consider, for instance the Buchwald–Hartwig reaction, which was first reported in 1994 \citep{Forero-Cortes2019-eh,Guram1994-yo, Paul1994-nz}.
To do this, we define a new test set of all the Buchwald–Hartwig reactions that are not present in any of the models' training sets (see Section~\ref{sect:timeBasedSplitsMethods} for more details), and then evaluate the models trained on each of the cutoffs on this new test set (\fig{fig:bhOverTime} and Table~\ref{tab:bh}). 
Note that in \fig{fig:bhOverTime}, the x-axis now represents the training set cutoff year  (which was previously represented by the different colored lines in  \fig{fig:timeSplits}).

From this analysis, we see that the earliest model (trained on reactions that were first reported up to 1996) starts with only one Buchwald–Hartwig reaction in its training set. (While the Buchwald–Hartwig reaction was first published in the academic literature in 1994, our models are trained on data extracted from the patent literature, and so even by 1996 very few Buchwald–Hartwig reactions had been used.) 
The model obtains a low top-1 accuracy of 5.2\% (8.2\% top-5) on the held-out Buchwald–Hartwig reaction test set, 
reflecting the lack of knowledge the model has on this reaction class at this point in time. 
When we look at the models trained on later cutoffs, we see that they have seen far more Buchwald–Hartwig reactions in their training sets, and this corresponds to far higher accuracies (i.e., top-1 accuracies above 60\%). 
Therefore, as expected, the ability to predict particular types of reactions depends on the availability of training data covering those types. Linking back to \fig{fig:timeSplits}, 
this difficulty that models have in extrapolating to new transforms can partly explain why accuracy falls off as the extrapolation distance increases. 
Nevertheless, while the model accuracy starts off low, it is not zero,
interestingly suggesting that some knowledge of this transform can also be inferred from the other reactions present.

\subsection{Reaction-type-based splits evaluate the strictest form of extrapolation across classes}
\label{sect:nameRxn}

If time-based splits serve partly as a surrogate for assessing performance on unseen reaction types, why not just evaluate on this task directly?
In fact, reaction discovery is likely not the only factor that occurs in a time-based split, as there are also shifts in the conditions used and the chemical space being explored (we come back to investigating the differences between our splits in \S\ref{sect:distributionShifts}).
Here, we instead assess the potential for successful generalization to new reaction types by using \emph{NameRxn splits}.

NameRxn is a rules-based, hierarchical method of classifying reactions \citep{NextMove-Software2022-ov,Lagerstedt2021-ct}. 
The hierarchy is inspired by and linked to other classification schemes \citep{Carey2006-ca,RSC2012-uo} and has been used for analyzing chemical trends \citep{Schneider2016-lq}.
The NameRxn hierarchy consists of three levels, the first dealing with high-level categories (e.g., classes include \nr{3. C-C bond formation}, \nr{4. heterocycle formation}, etc.), with the subsequent two levels then further filtering this down.
To illustrate, below \nr{3. C-C bond formation} the second level contains categories such as \nr{3.1 Suzuki reactions}, \nr{3.2 Heck reactions}, etc., and the third level below \nr{3.1 Suzuki reactions} contains the final classes \nr{3.1.1 Bromo Suzuki coupling}, \nr{3.1.2 Chloro Suzuki coupling}, etc.
The three digit code associated with each NameRxn in the bottom level (e.g., \nr{3.1.2} for \nr{Chloro Suzuki coupling}) indicates exactly where in the  hierarchy the NameRxn class lies.

Our reaction-type splits involve holding-out one or more of these NameRxn classes from the training set of our model to use as a test set. 
By doing this experiment for different NameRxn classes, we can get an idea of the ``chemistry'' the model can learn from other classes present in its training set
 and what forms of extrapolation it struggles with.
 Here we evaluate on five different reaction-type splits, holding out respectively, all (1) Grignard Ester, (2) Heck, (3) Chloro Suzuki, (4) Triflyloxy Suzuki, and (5) All Suzuki coupling reactions.
 Note that we consider three different classes of Suzuki splits: the Chloro Suzuki and Triflyloxy Suzuki splits consider subtypes of Suzuki reactions containing specific functional groups, whereas the All Suzuki split holds out all types of Suzuki reactions.  
 Further details on the exact NameRxn classes each split entails is provided in Section~\ref{sect:namerxnSplitsMethods}.

\begin{figure}
    \centering
    \includegraphics[width=\linewidth]{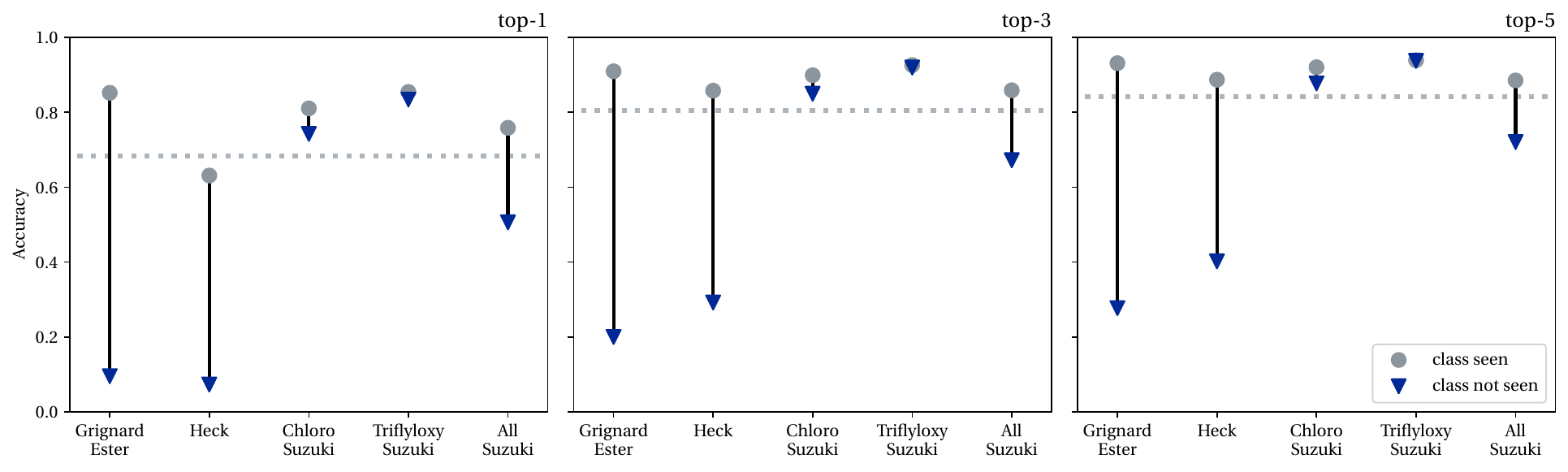}
    \caption[    Top-1,3,5 accuracies for reaction predictors evaluated on different reaction-type splits. ]{
    Top-1, 3, and 5 accuracies for reaction predictors evaluated on different reaction-type splits. 
    Each column shows the accuracy on a held-out set of a particular reaction class both (a) when seeing 1000 separate examples of the same reaction type during training (gray circles,
    \tikz{\draw[fill=grayRandom,draw=none] circle (.5ex);}; intrinsic difficulty) and (b) when seeing no reactions of that type during training (blue arrows, \tikz{\draw[fill=blueOOD,draw=none] (0,.5ex) -- (1ex,.5ex) -- (.5ex,-.5ex) -- cycle;}; extrapolation difficulty). 
    The gray dashed horizontal line shows the accuracy of a reaction predictor evaluated on an in-distribution test set (i.e., containing many different reaction classes).  
	Note that we remove all uncategorized reactions (NameRxn class \nr{0.0}) when creating our datasets.
	}
 % we use a short caption due to tikz incompatability -- https://tex.stackexchange.com/questions/56079/using-tikz-inside-a-figure-caption
    \label{fig:namerxnSplits}
\end{figure}

It is important to note that some reaction types can be \emph{intrinsically hard} to predict even when seeing other example reactions from the class, for instance due to containing complicated stereochemical transforms. 
However, our focus is on how hard the reaction types are to \emph{extrapolate to}, which is a distinct concept. 
To assess this for each NameRxn split, we divide the group of held-out reactions in two: 
we add the first subset (1000 reactions) to the training set for a baseline model, while reserving the remainder for testing only.
The baseline model's performance on the test set can give us an idea of the reaction type's intrinsic difficulty, and so puts the extrapolation performance in context (\fig{fig:namerxnSplits} and Table~\ref{tab:namerxn}).

Overall we see some small variation in the classes' intrinsic difficulties, with the predictions on the Heck reactions in particular being marginally less accurate than those for the other classes.
When it comes to extrapolation difficulty, the Suzuki splits, particularly the Chloro Suzuki and Triflyloxy Suzuki splits, appear the easiest. In contrast, the Grignard Ester and Heck reactions prove more challenging when reactions of those types are fully omitted from training.
Interestingly, even with these harder splits, the models still demonstrate some ability to extrapolate, particularly when we look at the top-3 and top-5 predictions.

\subsection{Deeper investigation into what enables reaction class extrapolation}
\label{sect:deeperInvestigation}

\begin{figure}[t!]
    \centering
\includegraphics[width=0.9\linewidth]{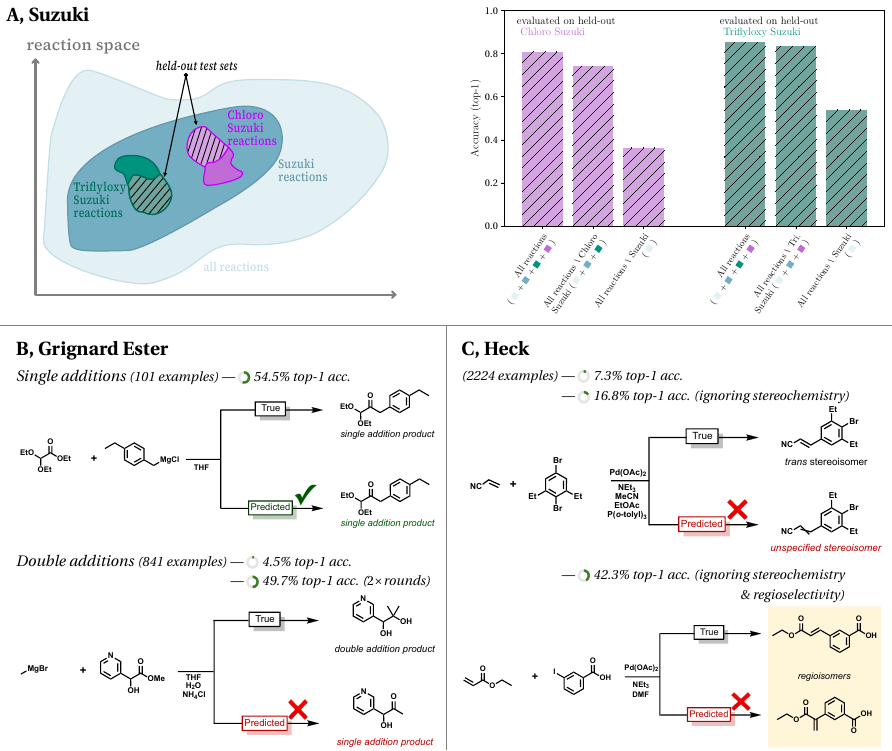}
    \begin{subcaptiongroup}
        \phantomcaption\label{fig:nameRxnExplanation:suz}
        \phantomcaption\label{fig:nameRxnExplanation:grigest}
        \phantomcaption\label{fig:nameRxnExplanation:heck}
        
    \end{subcaptiongroup}
    \captionsetup{subrefformat=parens}
    \caption{
    We investigate reasons for the contrasting  performance in the different NameRxn splits.
    \textbf{\subref{fig:nameRxnExplanation:suz}} For the Chloro Suzuki and Triflyloxy Suzuki splits, we assess whether the large number of other Suzuki reactions present can explain the good extrapolative performance.
    Namely, to evaluate on our specific  Chloro and Triflyloxy Suzuki test sets, we create three different training sets (as shown by cartoon, left): (i) reactions from all classes (including separate reactions from the same specific Suzuki class); (ii) reactions only from other specific Suzuki reaction classes (and non-Suzuki reactions); and (iii) non-Suzuki reactions only. 
    Results are shown on the right. The different bars show the accuracy for the different cases (the square color boxes in the x-axis labels indicate the reaction classes used in training the respective models).
    \textbf{\subref{fig:nameRxnExplanation:grigest}}For the Grignard Ester split we notice that the model does particularly poorly on a \emph{double addition} subset, but such a reaction can actually be expressed as two single additions and that when we allow our model to do two rounds of predictions (i.e., when we feed the predicted product from the first round in as an input the second time around) performance improves.
    \textbf{\subref{fig:nameRxnExplanation:heck}}
    For the Heck split we see that the model particularly struggles with the stereochemistry and regioselectivity present in these reactions (see text for further details).
    }
    \label{fig:nameRxnExplanation}
\end{figure}

We can further investigate how the model is able to extrapolate in these splits, and conversely why sometimes it is unable to do so.
We look first at the Chloro and Triflyloxy Suzuki splits.
Remarkably the model finds it easy to extrapolate to these classes: the drop in top-1 accuracy is less than 10\%.
We hypothesize that this is in part due to the large number of other Suzuki reactions that remain in the model's training set, and so we investigate how excluding these reactions (using our All Suzuki split) affects results (see \fig{fig:nameRxnExplanation:suz}).
\fig{fig:nameRxnExplanation:suz} confirms that removing these other Suzuki reactions has a far more pronounced effect on the extrapolation difficulty than removing reactions in the specific subclass, suggesting that the model does indeed learn from the Suzuki class as a whole. 
Having said that, even in this more challenging extrapolation, accuracy does not drop to zero, which could be attributed to the existence of many structurally similar non-Suzuki cross-coupling reactions, such as the Kumada coupling, Negishi coupling, Stille coupling, and others.

Secondly, analyzing the Grignard Ester reactions, we find that they can be divided into two main groups: single and double additions (see \fig{fig:nameRxnExplanation:grigest}). Our model particularly struggles to predict the latter of these, exhibiting a top-1 accuracy of 4.5\% on the double additions compared to the baseline model's 91.1\% (on the single additions both the model and the baseline get around 55\%).
This can be explained because most of the dataset's remaining (i.e., non-ester) Grignard reactions form single-addition products via the protonation of a stable tetrahedral magnesium alkoxide intermediate. 
In the Grignard Ester examples however,  this species collapses to give a carbonyl group that is often more reactive than the ester starting material and thus (unless burdened by steric effects) necessarily reacts again with the Grignard nucleophile, most often yielding the double addition product. Thus, double additions can be mechanistically described as the continuation and repetition of a single addition reaction, and one therefore could model them as the composition of two single addition reactions. We reframed this task accordingly by feeding the product from the initial prediction back into the model as a reactant. With this change, the model improves from 4.5\% to 49.7\% top-1 accuracy on the double additions, showing that the latent chemical knowledge needed to predict Grignard Ester products already exists in the model, but the knowledge of what should be considered a \emph{single} reaction step is not.

In the case of the Heck reaction split, many incorrect predictions are a result of the multiple possible isomeric products this reaction can produce: \textit{E-} or \textit{Z-}stereochemistry occurs due to the double bond formed, while different regioisomeric products can also result depending on factors such as the type of catalyst. 
To quantify how well the model deals with such nuances, we first re-evaluate the top-1 accuracy after removing all stereochemical information from both the predicted and ground-truth products.
When doing so, we find that both the Heck-withheld and baseline models' accuracies improve (from 7\% to 17\% and 63\% to 85\%, respectively), suggesting that stereochemistry may be intrinsically harder to predict rather than merely challenging to extrapolate to (which could clarify the baseline model's relative performance in \fig{fig:namerxnSplits}).

Extending this analysis to regiochemistry, the top-1 accuracy of the Heck-withheld model further increases to 42\% when different regioisomers are counted as correct predictions (\fig{fig:nameRxnExplanation:heck}), while the baseline model's accuracy increases to 89\%.
Overall this confirms that the model's performance on the Heck reaction can in part be attributed to the wide range of possible Heck products, and that this challenge is significantly amplified in extrapolative settings, where trends in regioselectivity in particular are hard to model without seeing the reaction class.

\subsection{Analysis of distribution shifts offers insights into the relationship between split types}
\label{sect:distributionShifts}

In the analysis done so far, we have taken an \emph{objective-driven} approach to discussing and evaluating reaction extrapolation. 
That is, if your objective is to better understand how reaction predictors work on already explored reaction spaces, then use a document- or author-based split;
if your objective is to understand how reaction predictors work prospectively, then use a time-based split;
or if your objective is to understand how well reaction predictors can generalize to new types of transforms, then use a reaction-type split.

However, this is not the only approach one can take when designing out-of-distribution tasks.
An alternative would be to take what we would call a \emph{data-driven} approach to split design (we discuss relevant related work below and also in the next section). 
In particular, this looks at the data given to the model---typically focusing on the inputs---and picks a particular feature of this data to split on that ideally is not important for prediction; in certain cases it might even generate synthetic out-of-distribution data by editing existing data in a way that should not affect the label.
Examples of data-driven approaches include adding Gaussian blur to images on classification tasks \citep{Hendrycks2017-ie}, splitting on molecular structure in molecular classification tasks \citep{Koh2021-no}, or splitting by molecular weight in reaction prediction tasks \citep{Gil2023-rm}.
Taken to the extreme, adversarial examples \citep{Goodfellow2014-gq,Szegedy2013-kr,Carlini2019-da,Bao2022-nr} are out-of-distribution data specifically designed from the input data to be as challenging as possible to the model.
Although the resultant splits from a data-driven approach can end up testing similar properties of the model, the approach differs in how the splits are derived.

\begin{figure}[t]
    \centering
    \includegraphics[width=1.\linewidth,trim= 0 10 0 0, clip]{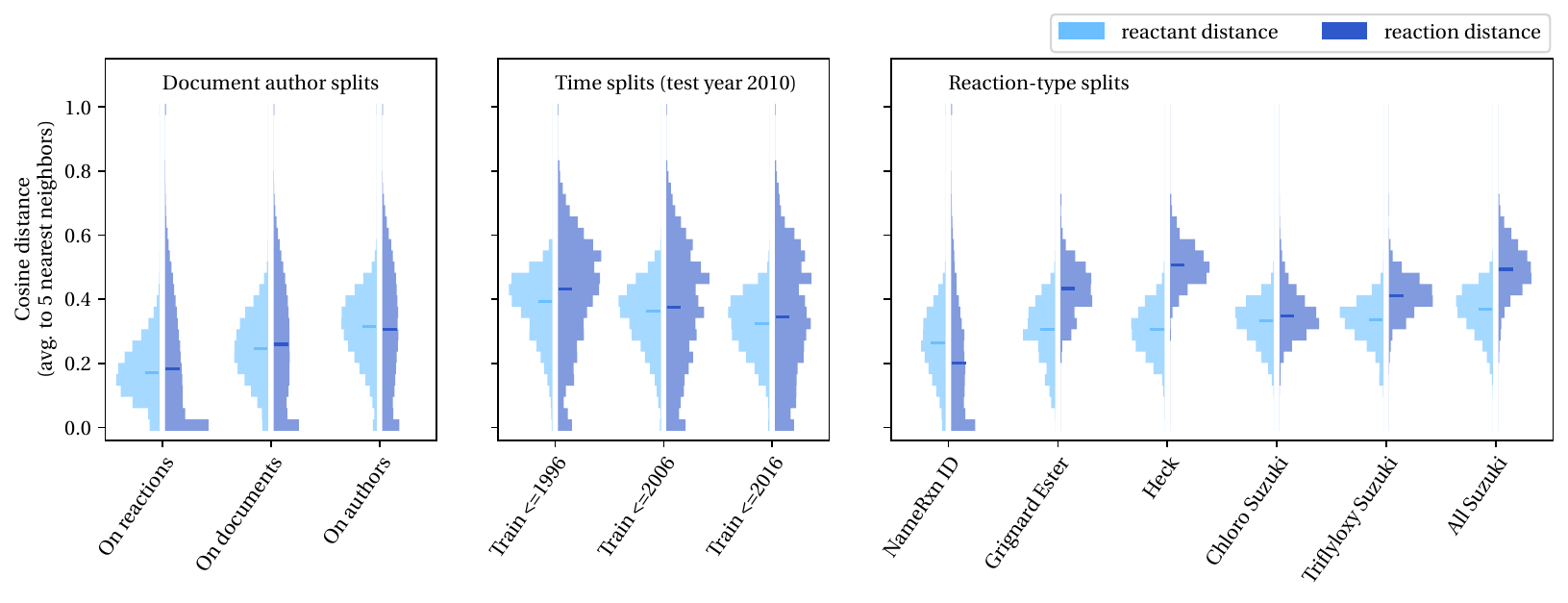}
    \caption{Distribution of the average cosine distance between each reaction in each test set to its nearest five neighbors in the corresponding training set. The left-hand part of each histogram (in the \textcolor[HTML]{A6D9FF}{lighter blue}) shows the distance in reactant fingerprint space (i.e., it indicates how different the test reactant molecules are to those the model sees in training), whereas  the right-hand part of the histogram (in the \textcolor[HTML]{829BDF}{darker blue}) shows the distance in reaction fingerprint space (i.e., it indicates how different the reaction transform is to those seen in training). The solid horizontal lines indicate the median of each distribution. Further details about splits and how fingerprints are calculated can be found in Section~\ref{sect:violingPlot}.  }
    \label{fig:violinReactantVersusReaction}
    \vspace{-0.4cm}
\end{figure}

It is worth reassessing the splits we have considered from this point of view and comparing them in terms of how the input data differ between the training and test sets.
When looking at the input to a reaction predictor, one can change it in two main ways. 
One can either consider (a) varying the reactants, i.e., consider performing the same reaction over different areas of reactant space; or (b) varying the transform, i.e., consider performing very different reactions over similar areas of reactant space.
In \fig{fig:violinReactantVersusReaction} we visualize how the splits vary along these two dimensions by plotting the distribution of distances from each test point to its five nearest neighbors in the training set in terms of (a) reactant fingerprint distances and (b) reaction fingerprint distances. 
While structural fingerprints do not provide a complete picture of how reactions may differ along these different dimensions, they are still useful as a quick-to-compute and practical guide.

\fig{fig:violinReactantVersusReaction} shows that the distance to the nearest neighbors increases in both reactant and reaction fingerprint space on our out-of-distribution datasets, as one might expect. 
For the document- and author-based splits, we see a large shift in the reactant distance distribution (particularly when looking at the medians) and also a drop in the number of neighbors at zero distance in reaction space.
Many patents contain multiple examples of one reaction type applied to many structurally-similar reactants. 
The changes seen in distributions match what we might expect as we prevent the model from training and testing within these same substrate scopes, to training and then testing on different ones, highlighting what makes these splits more challenging.

The distributions for the time-based splits demonstrate that both the reactants and reactions change together as the model extrapolates further into the future. 
This differs from the reaction-type splits, where the distance in reaction space is greater than in reactant space, reflecting the design of these splits.
Interestingly, the distance also seems somewhat correlated with the observed empirical difficulty of the split, with further distances seen more often for the harder Grignard Ester, Heck, and All Suzuki splits compared to the easier Chloro and Triflyloxy Suzuki splits.
Therefore, such distances could be used to design further challenging splits.

\section{Discussion} % for asides / footnotes from earlier?
\paragraph{OOD benchmarks enable OOD improvement.} The importance of considering the data generating process and evaluating on OOD data has long been prevalent in ML more generally (see e.g.,  \citep[Fig. 5]{LeCun1990-jn}, or \citep{Quinonero-Candela2008-fp, Sugiyama2012-mm, Scholkopf2012-dq, Peters2017-ff}), with related developments including both better benchmarks \citep{Koh2021-no,Santurkar2021-qw,Yang2022-en,Srivastava2023-qt, Gulrajani2020-sp} and modeling/algorithm advancements to better deal with such data \citep{Shimodaira2000-it, Hendrycks2017-ie, Lakshminarayanan2017-mo, Hu2022-iq}. 
Within the domain of ML for chemistry, OOD splits---such as scaffold, time, and others---have facilitated the development of more robust models for molecular regression and classification tasks \citep{Wu2018-ii,Sheridan2013-ys, Landrum2023-on, Kanakala2023-tk,Tossou2024-ew, Steshin2023-jm}.

\paragraph{OOD benchmarks in reaction prediction.}
When it comes to reaction prediction, many models are introduced then tested only on in-distribution splits. Performance is sometimes broken down into how well methods do on subclasses of reactions (see, e.g., \citep[Table 5]{Schwaller2019-ra} or \citep[Fig. 6]{Tetko2020-yb}), but this is carried out in situations where the models have seen the same classes in their training set (i.e., an in-distribution setting). Sometimes single time-based \citep{Segler2018-ns}, template \citep{Seidl2022-do}, or document \citep{Tu2022-xr} splits have been used, but this practice is not widespread. Alternative metrics to top-1 accuracy to evaluate methods have been studied in the context of in-distribution settings \citep{Jaume-Santero2023-aj} as well as works developing and discussing how to better deal with noisy reaction data \citep{Griffiths2018-xb,Toniato2021-dy,Andronov2023-ox}.

When OOD tasks have been investigated in reaction prediction, this is often in the context of method development for enabling models to adapt to new reaction types or different, often proprietary datasets \citep{Toniato2023-ac,Su2022-xh,Wang2020-nn,Wang2022-ws,Pesciullesi2020-ma}.
Techniques range from multitask learning to fine-tuning, but on the whole, evaluation is carried out on only a single type of split, often representing single reaction classes (for instance, \citet{Wang2020-nn} focuses on the Heck reaction and \citet{Su2022-xh} looks at the Chan–Lam coupling and how this relates to other, similar reaction classes). 

Perhaps closer to the ideas here, there has also been a series of recent work reappraising how well existing reaction predictors models work: both in the forward \citep{Gil2023-rm,Kovacs2021-bh,Wigh2024-no} (like our work here) and the inverse direction \citep{Chen2024-lk,Yu2024-pr}.
For instance, \citet{Gil2023-rm} proposed a new open-source benchmark for reaction prediction that includes a variety of  new metrics, such as OOD accuracy on molecular weight–based splits as well as sustainability-related factors such as \ce{CO2} emissions.
Elsewhere, \citet{Kovacs2021-bh} developed methods to relate model predictions to both their current inputs and previously seen training data, using this to motivate new template- and scaffold-based splits.
Focusing on the inverse direction (i.e., retrosynthesis) instead, \citet{Yu2024-pr} developed OOD template- and size/scaffold-based splits to evaluate models on different forms of extrapolation; we discussed how the splits we consider can be viewed in a similar framework in Section~\ref{sect:distributionShifts}.
While these works share much of our philosophy---of trying to better characterize current reaction predictors' already existing generalization abilities---they do not consider the breadth of different OOD tasks we consider here. 
Therefore, working out how to incorporate our splits into new benchmarks, such as that proposed by \citet{Gil2023-rm}, would make an interesting future direction.

\paragraph{Interplay between model and data.} 
We have argued that a key factor in designing better evaluations is to consider the provenance of the data and its structure. 
While here this has involved considering the human aspect of dataset curation (\S\ref{sect:docAuth}), further down the line an optimal prospective study might introduce how a ML-based model additionally interacts with the data-generating process \citep{Kearnes2021-dt}. 
In considering such a benchmark there will always be trade-offs to be made between the faithfulness of the setup versus the practicalities of being able to run it.

\paragraph{Splits are useful in unison.} Ultimately, there is not a single ``best'' type of split to use when evaluating reaction predictors. 
In Section~\ref{sect:distributionShifts}, we analyzed how the splits differed and discussed how they can be linked to particular modeling objectives.
However, even within a single objective it is important to consider multiple splits.
For instance, when evaluating if a model can correctly ``discover'' a new reaction,
neither time-based splits or reaction-type splits capture the full picture alone:
time-based splits also represent other shifts such as substrate changes, while reaction-type splits ignore aspects of the real causal discovery process. (For instance, imagine that the reaction discovery process goes reaction types $A \rightarrow B \rightarrow C$, and that we are holding out and evaluating on reaction type $B$. Then the model can use both the information from reaction types $A$ and $C$ when making this prediction, even though the latter information was not available during  reaction type $B$'s real discovery.) 
Therefore, different splits offer complementary information about model capabilities and it is useful to use them together.

\section{Conclusion}

In this work, we have highlighted the over-optimistic nature of current reaction predictor evaluations due to their unrealistic in-distribution setting, and discussed and evaluated ways to build better benchmarks.
We have shown how document-based splits better account for the inherent factored structure of existing datasets, time-based splits enable prospective evaluation of a model's performance, and reaction-type splits can assess a model's capability for reaction discovery.
By providing more faithful measures of current reaction predictor performance for different use cases, we draw attention to areas where they can be improved, to enable these shortcomings to be addressed by next-generation models. 

The benchmarks we propose can also be further improved and developed. 
For instance, we only consider patent data here, which may have some key differences to datasets derived from academic papers or high throughput experimentation. 
Other future directions to pursue include exploring additional reaction-type splits, investigating how models might adapt quickly (e.g., using in-context learning), as well as analyzing how modeling choices might affect generalization.

Looking forward, we view reaction discovery as one of the most appealing use cases for reaction predictors. Although additional techniques are required to build a practical system (e.g., a method of sampling which molecules to feed through to a predictor), our more representative evaluation settings---and the insights they deliver---act as an important component of progressing towards this goal.

\section*{Acknowledgments}
This work was supported by the Machine Learning for Pharmaceutical Discovery and Synthesis consortium and the National Science Foundation under Grant No. CHE-2144153 (to CWC). Additional computational resources were also provided by the MIT SuperCloud and Lincoln Laboratory Supercomputing Center. AZ received additional support from the NSERC PGS-D fellowship. BM received additional support from the MIT-Novo Nordisk Artificial Intelligence Postdoctoral Fellows Program.

\bibliography{refs}  

% make supplementary material have S headings here so that distinct from sections in the main text:
\clearpage 
\begin{appendices}
\renewcommand{\thesection}{S\arabic{section}}
\renewcommand{\thesubsection}{\thesection.\arabic{subsection}}
\setcounter{figure}{0}
\setcounter{table}{0}
\renewcommand{\thefigure}{\thesection.\arabic{figure}}
\renewcommand{\thetable}{\thesection.\arabic{table}}
\section{Methods}
\FloatBarrier
\label{sect:methods}

In this section we describe further details about the data and models we use.
Specifically, Section~\ref{sect:dataset} describes the dataset used and how this is pre-processed, while Sections~\ref{sect:docMethods}–\ref{sect:namerxnSplitsMethods} describe the individual splits. Section~\ref{sect:modelMethods} describes how we tune, train, and evaluate our models, and Section~\ref{sect:violingPlot} provides further details of how plots in the main text were created.
Further methodological details can be found in our code, available online at:
\begin{description}
    \item[\splitCode:] for cleaning the datatset and creating our splits (i.e., the processes described in \S\ref{sect:dataset}–\ref{sect:namerxnSplitsMethods}).
    \item[\trainingCode:] for training and evaluating our models (\S\ref{sect:modelMethods}).
\end{description}

\subsection{Creating a clean reaction dataset}
\label{sect:dataset}

Our dataset and splits are created from the 2022Q4 version of the Pistachio dataset \citep{Mayfield2017-qy,NextMove-Software2021-td} (although not presented here, we also performed similar experiments with earlier versions of this dataset, finding similar qualitative trends). 
We restrict ourselves to the \emph{US grants} part of this dataset (as opposed to for instance the \emph{applications} part) to limit the same reactions coming up from different jurisdictions and filings.
To create a cleaned dataset from these reactions, we  perform three steps: (1) standardization, (2) filtering, and (3) deduplication, the details of which we go into below.
While these steps likely do not remove all incorrectly reported reactions, they weed out a number of strange reactions and ensure that reactions that are completely identical cannot occur in both the training and test sets.
We leave investigations into extending these steps and better dealing with missing ground truth details (e.g., additional products created) as future directions to explore.

\paragraph{(1) Standardization.} To standardize reactions we remove any atoms maps and canonicalize the molecules using RDKit \citep{RDKit-Team2021-tp} (if the canonicalization fails then we skip the reaction). 
We also remove any ChemAxon SMILES extensions,\footnote{See \url{https://docs.chemaxon.com/display/docs/formats_chemaxon-extended-smiles-and-smarts-cxsmiles-and-cxsmarts.md}} but keep any stereochemical information that is encoded directly in the SMILES (e.g., using the symbols `\texttt{$\backslash$}', `\texttt{@}', etc).
Reagents (i.e., molecules present but not contributing heavy atoms to our products) are mixed with reactants, and in general we make no distinction between these and ordinary reactants in our experiments.

\paragraph{(2) Filtering.} Going through our standardized reactions, we then filter out any reactions that do not meet a certain set of criteria. 
The aim of these criteria are to identify both strange reactions (e.g., those involving very large molecules) and non-interesting reactions (e.g., those that only describe the disappearance of a reactant), which complicate any downstream analysis. 
Specifically, we remove any reactions for which any of the following conditions are true:
\begin{enumerate}
	\item the reactants have fewer than 5 heavy atoms, 
	\item the reactants contain no carbon atoms, 
	\item none of the reactants have at least two bonds, 
	\item the reaction is easily identifiable as a (de)protonation (we neutralize commonly occurring charged atoms in the reactants and products using a SMARTS pattern and, having done so, see if  the sets of reactants and products are then equal), 
	\item all products not already present in the reactant set contain fewer than 2 heavy atoms, 
	\item the reaction is very long when tokenized\footnote{Details on the tokenizer we use can be found in \S\ref{sect:modelMethods}.} for our language model–based reaction predictor (e.g., over 800 tokens long).
\end{enumerate}

\paragraph{(3) Deduplication.} Finally, we deduplicate reactions by putting each reaction's reactant(s)-product(s) pair into a canonical representation.
When picking between duplicates, we typically keep the first one we encounter when iterating through the Pistachio dataset;
however, we displace the first one if it is missing a NameRxn tag (and the subsequent reaction we encounter is not) or if, failing this, the year associated with the subsequent reaction is earlier than the year associated with the first reaction we encountered.
This is so that we try to keep the earliest recorded complete occurrence of each reaction.
While this deduplication routine is not perfect, it ensures that at test time  we do not evaluate on an example that completely matches one in the training set.

\begin{figure}[t]
    \centering
    \includegraphics[width=0.8\linewidth]{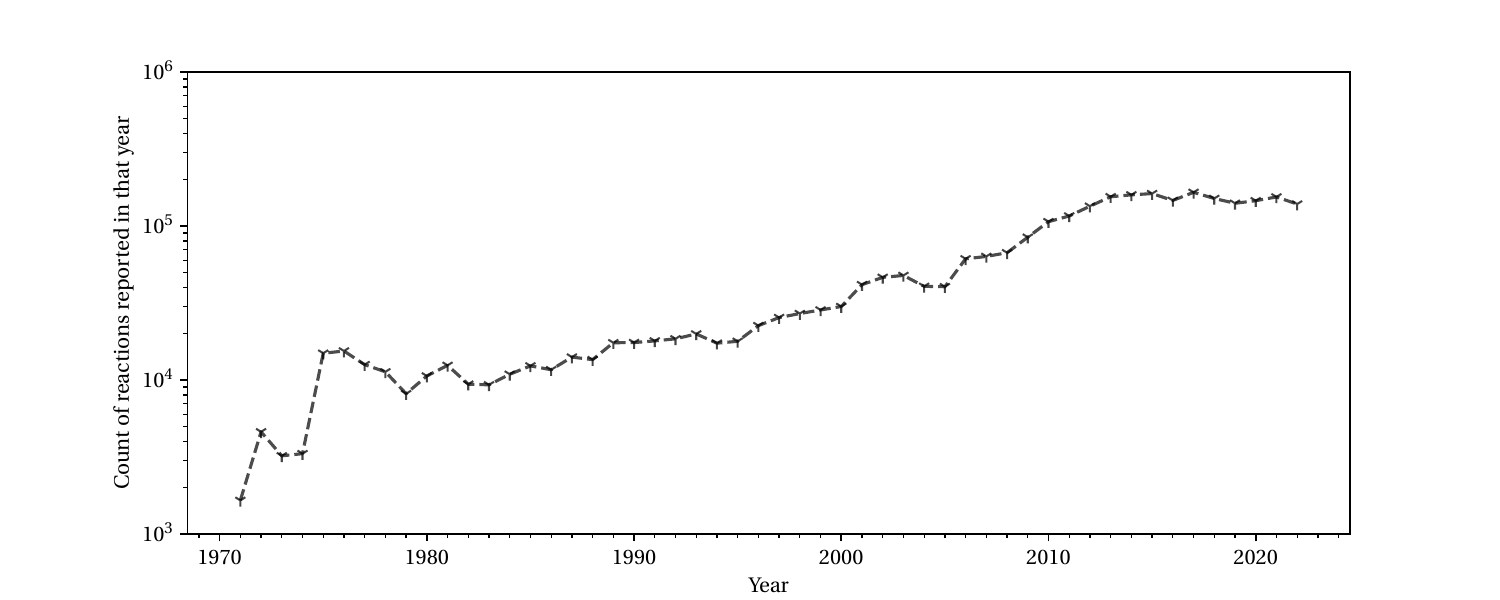}
    \caption{
    	The number of reactions associated with each year in our cleaned data from 1971–2022.
    	We refer interested readers to the works of  \citet{Schneider2016-lq,Roughley2011-pi} for further details on how the distribution of reactions recorded in typical datasets has evolved over time.
	}
    \label{fig:reactionCounts}
\end{figure}

\paragraph{} In total we end up with just over 2.8mn (million) unique reactions. These are used as a starting point for creating our individual splits.
Overall, these reactions come from over 200k (thousand) different documents (where each document is distinguished by the first part of the patent number), representing the work of just under 180k different authors spread across nearly 11k different assignees.
The number of reactions associated with each year in our final deduplicated dataset is presented in \fig{fig:reactionCounts}.
One can see that generally the number of reactions reported per year increases over time.

\subsection{Document- and author-based splits}
\label{sect:docMethods}

In our document- and author-based splits we create an ID training set size of 1mn reactions, and three test sets of 100k reactions each: an ID and two OOD test sets (we also create an ID validation set for hyperparameter tuning containing 30k reactions). 
This is done in a series of steps, the key parts of which are detailed below. 
Note that we use the document title (as it exists in Pistachio) when referring to documents, but our process could also be extended in the future to tie together related documents, for instance by using patent citation information.

\paragraph{(1) Defining author-to-document and document-to-reaction maps.} 
First, we take our cleaned, processed dataset (described in the previous section) and define a mapping from authors to documents (a many-to-many relationship) and a mapping from documents to reactions (a one-to-many relationship due to the deduplication process described in Section~\ref{sect:dataset}).

\paragraph{(2) Splitting on authors.}
Using our constructed mappings, we then create our author-based split. 
To do this we iterate through our list of possible authors (in a random order) and assign all the associated documents (if they have not been encountered already) first to an OOD author set until that is full (i.e., contains more than 100k reactions to create the author-based test set) and then to an ID author set until that is full (i.e., contains more than 1230k reactions to create our remaining training and test sets---described next).

\paragraph{(3) Splitting on documents.} 
The documents associated with the ID author set are further divided into two.
This happens in a similar manner to how the authors were separated: we iterate through our list of documents (in a random order) and assign all the associated reactions first to an OOD document set until that is full (i.e., contains more than 100k reactions to create the document-based test set) and then to an ID document set until that is full (i.e., contains more than 1130k reactions to create our ID sets).
The reactions in the ID document set are randomly divided into the ID sets: 1mn for the training set, 100k for the ID test set, and 30k for an ID validation set (used for hyperparameter tuning).
\paragraph{}
The end result of this, as we mentioned above, is three separate test sets, each 100k reactions in size (one split \emph{on authors}, one \emph{on documents}, and one only \emph{on reactions}), as well as a 1mn reactions training set and a 30k reactions validation set. In total, we use 1.33mn reactions from our cleaned dataset (47\% of the total) to form all the training/validation/test sets used in the author- and document-based splits; these reactions represent the work of just over 100k authors (56\% of the total) and just over 72k documents (36\% of the total).

\begin{figure}[t]
    \centering
    \includegraphics[width=\linewidth]{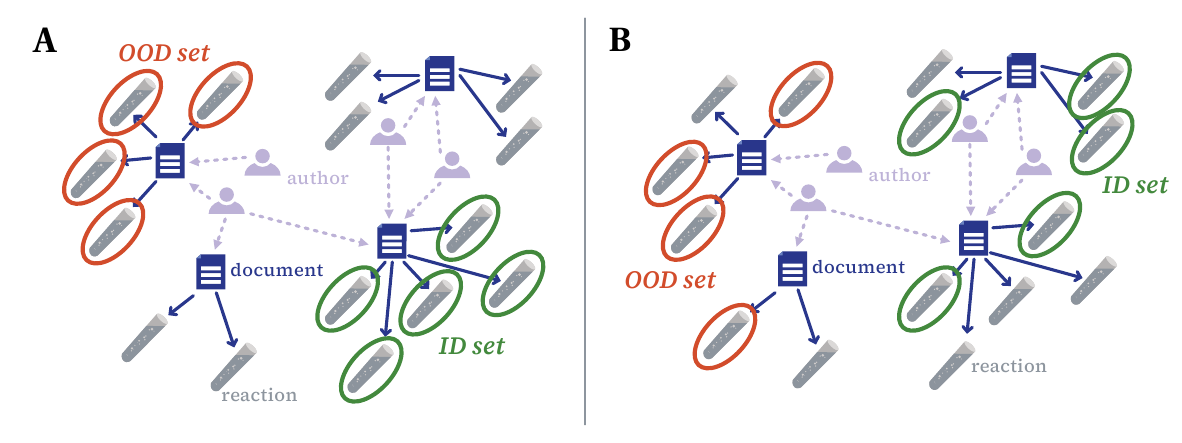}
    \caption{
        The splitting procedure we use is more likely to form a document-dense split (panel A), where
        the OOD set (shown in \textcolor{oodRed}{red}) and ID set (shown in \textcolor{idGreen}{green}) are sampled from  a small number of documents---see text for further details.
        While other document-based splits could also be considered (such as a split similar to that shown in panel B), our
        splitting procedure has the advantages of being data efficient and also reflecting the author-document-reaction structure found in the complete dataset (i.e., without any subsampling).
	}
    \label{fig:docDenseCartoon}
\end{figure}

Note that our splitting procedure means that the datasets are subtly different from subsampled reaction sets considered elsewhere: they are ``author-document-dense.''
By ``author-document-dense,'' we mean that when forming the sets, we try to use the minimum possible number of authors and documents necessary. 
In other words, if one reaction from a document ends up in our OOD document test set, then all the other reactions associated with that document are \emph{very likely}\footnote{
We use the term \emph{very likely} as some reactions will be discarded from the list of OOD reactions to get back down to a 100k test set when finalizing the datasets.
} to as well (\fig{fig:docDenseCartoon}).
Ultimately, this may mean that our sets represent a smaller total area of chemical space than from a random subsample.\footnote{In a random subsample one could model the counts of reactions taken from each document as following a multivariate hypergeometric distribution.} 
This observation highlights a central argument of our work: when considering the performance of reaction predictors, it is not enough to simply consider the size of the datasets they are trained on—it is critical to bear in mind their provenance too.

\subsection{Time-based splits}
\label{sect:timeBasedSplitsMethods}

The time-based splits (\S\ref{sect:timeSplits}) are created from our cleaned dataset by first ordering all reactions by year.
Starting at 1976, we break off a held-out test set of 3k reactions for each year. 
(When creating these splits we use a document-based splitting strategy---we explain why below.)
The remaining reactions are used to create a separate training and validation set for each cutoff; 
each of these contains reactions that were reported up to (including in) the designated year.
We create the training/validation sets for each cutoff independently (i.e., the reactions in the training set for an earlier cutoff do not influence the reactions chosen for the training set in a later cutoff). 

When producing the time-based split results reported in the main paper, we control for dataset size. 
The validation sets each contain 2k reactions, and the training set sizes are just under 250k reactions each---%exact number is 248916 I think.
this is the maximum available reactions we could use in the earliest split.
In Section~\ref{sect:moreTimeSplits}, we examine time-based splits that do not control for training set size (the validation set still comprises 2k reactions); here, the training set size is still just under 250k reactions for the 1996 cutoff set (this represents all that are available), but rises to approximately 2.26mn reactions for the final 2020 cutoff set. 

\paragraph{Using a document-based splitting strategy when creating test sets.}
As mentioned above, when creating the test sets for each year, we use a document-based splitting strategy.
This is because a time-based split implicitly creates document-type splits after the model's cutoff point (due to each document being associated with only one publication year).
Without a document-based spitting strategy elsewhere, this transition manifests as a sharp drop in accuracy at the cutoff point.  
This drop complicates our primary analysis into investigating the extrapolation difficulty due to distribution shifts over time.
Therefore, to maintain consistency, we employ a document-based splitting strategy when creating the held-out test set for every year.

\paragraph{Creating the Buchwald–Hartwig test set.}
To create the Buchwald-Hartwig test set (\S\ref{sect:buchwaldHartwigAnalysis}), we go back through our cleaned dataset and extract all of the Buchwald-Hartwig reactions that were not included in our previously created training and validation sets. 
This process results in approximately 16k Buchwald-Hartwig reactions in total, corresponding to the NameRxn codes \nr{1.3.1}, \nr{1.3.2}, \nr{1.3.3}, \nr{1.3.4}, and \nr{1.9.43}.

\subsection{Reaction-type splits}
\label{sect:namerxnSplitsMethods}

As outlined in the main text, we created the reaction-type splits (\S\ref{sect:nameRxn}) using the NameRxn classification system.
The NameRxn code for each reaction is provided as part of the Pistachio dataset. 
To create our splits using these, we first removed all uncategorized reactions (with NameRxn code \nr{0.0}) from our cleaned dataset, to avoid inadvertently training on reactions that might be similar to those that we are trying to exclude (note that this means that the accuracy results for our reaction-type splits are not directly comparable to the other splits considered).
The remaining reactions are then split independently six times: five times with our different held-out reaction types and once holding out no reaction types at all (this ``base'' split is used for hyperparameter tuning).
The following NameRxn classes are used for each split:
\begin{description}
	\item[Grignard Ester:] \nr{3.7.14 Bromo Grignard + ester reaction}, \nr{3.7.15 Chloro Grignard + ester reaction}, \nr{3.7.17 Iodo Grignard + ester reaction}, and \nr{3.7.19 Grignard ester substitution}.
	\item[Heck: ] \nr{3.2 Heck reaction} (including all subclasses).
	\item[Chloro Suzuki:] \nr{3.1.2 Chloro Suzuki coupling} and \nr{3.1.6 Chloro Suzuki-type coupling}.
	\item[Triflyloxy Suzuki:] \nr{3.1.4 Triflyloxy Suzuki coupling} and \nr{3.1.8 Triflyloxy Suzuki-type coupling}. 
	\item[All Suzuki:] \nr{3.1 Suzuki coupling} (including all subclasses).
\end{description}

Each split creates five different sets: a 1mn reactions training set, a 10k reactions validation set, a 10k reactions ID test set, and two OOD reaction sets made up of only reactions from the held-out reaction classes.
The first OOD set contains 1000 reactions and is added to the training set when training the baseline model for that split (see main text for details);
the second, containing what remaining OOD reactions are available (and capped at a maximum 10k in size), is used as our OOD test set.
As a result mainly of the amounts of different reactions available in our cleaned dataset, the OOD test set is approximately 1k reactions in size for the Grignard Ester split, 2k reactions in size for the Heck split, 3k in size for the Triflyloxy Suzuki split, and 10k in size for the other splits.
(We emphasize that only the test set size differs for the different splits, the training set size is kept constant.) %
% 974 grignard ester, 2224 heck, 2694 triflyoxy suzuki
When creating the different reaction-type splits, we use a document-based splitting strategy for consistency with the earlier experiments.

The NameRxn system is helpful for our use case due to its good coverage (74\% of our cleaned dataset has been categorized) and hierarchical nature (meaning we can split at different levels).
However,  it is not the only classification system suitable for assessing the ability of reaction predictors to generalize to new reaction types. 
Different classification systems will likely have different classification rules and different relationships between the classes they consider, leading to other advantages and disadvantages, and so could be an interesting future direction to explore.

\paragraph{Identifying single and double additions in the Grignard Ester split.} 
For the analysis in Section~\ref{sect:deeperInvestigation} we further split the Grignard Ester set down into single and double addition subsets. 
This was done by writing out a SMARTS pattern to count the number of alcohol groups present in the products versus the reactants. 
While  such an approach may produce a few false positives, we found this practical method simple and generally effective.

\subsection{Model}
\label{sect:modelMethods}

\begin{figure}[t]
\centering
\includegraphics[width=0.8\textwidth]{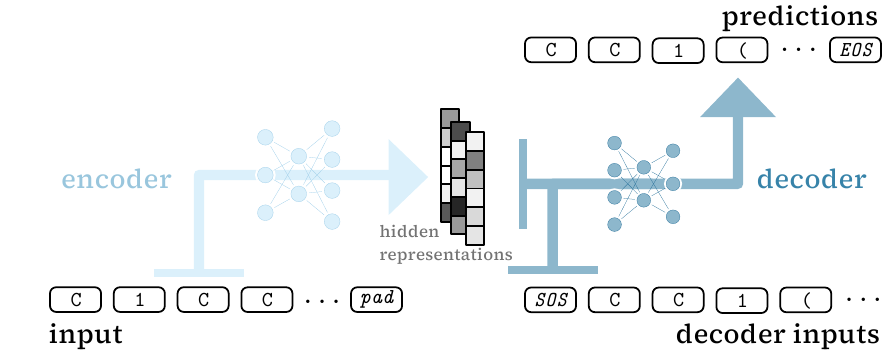}
\caption{We use an encoder-decoder Transformer model  for reaction prediction based on the BART architecture \citep{Lewis2020-nf}. 
SMILES are tokenized using the scheme proposed by \citet{Schwaller2018-rh}  and special tokens (e.g., for \emph{padding}, \emph{start of sequence}, and \emph{end of sequence} are added as appropriate).
Both the encoder and decoder use attention mechanisms to focus on different parts of the initial input (i.e., the reactants) and decoder inputs (i.e., the parts of the products predicted so far).
}
\label{fig:bart}
\end{figure}

For the experiments, we use an encoder-decoder Transformer model \citep{Vaswani2017-se} based on the BART architecture \citep{Lewis2020-nf} (\fig{fig:bart}).
Specifically, we use the  implementation from HuggingFace's Transformer library \citep{Wolf2019-tf}, but with the SMILES tokenization scheme\footnote{We modify this scheme slightly to extend it to cover the large loop numbers that can occur in the SMILES of the Pistachio dataset.}
proposed by \citet[\S3.1]{Schwaller2018-rh} as opposed to byte-pair encoding. (Generally we expect the difference in tokenization to have a minor effect; see \citet[\S4.1]{Chithrananda2020-dn} for a discussion on the differences in molecular tokenization performance for a molecule representation task).
We do not pretrain our model on a denoising task, instead training from scratch for each task in a supervised manner only. 
Overall the model used is very similar to \citet{Schwaller2018-rh}'s Molecular Transformer.

\subsubsection*{Training and evaluation}

\begin{table}[t]
    \centering
    
    \caption{Grid used for hyperparameter tuning on the Pistachio dataset splits.
    Please see our code (\trainingCode) for further details.}

    \small
    \begin{tabularx}{\textwidth}{XXp{0.4\textwidth}}

    \toprule
    \normalsize{\textbf{Hyperparameter}}  & \normalsize{\textbf{Grid}}  & \normalsize{\textbf{Note}} \\
    
    \midrule

    Gradient accumulation steps & $\{2, 4, 8, 16\}$& We use a fixed batch size and control the effective batch size using gradient accumulation (so that larger batches can fit in GPU memory).\\
    \addlinespace
    
    Learning rate & $\textsf{loguniform}(1e-5, 1e-2)$ & We use the AdamW optimizer \citep{Loshchilov2019-wh}. \\
    \addlinespace

    Warmup steps  & $[100, 10000]$ & We use a cosine scheduler with linear warmup.  \\ \addlinespace

    Encoder layers & $\{2, 3, 4, \ldots, 12 \}$ & \\
    \addlinespace

    Encoder FFN dim. & $\{ 512, 1024, 2048, 4096 \}$ & The dimension of the intermediate layer for the feedforward network in the encoder. \\
    \addlinespace

    Encoder attention heads  &  $\{ 4, 8, 16, 32 \}$  & Number of attention heads in the encoder. \\
    \addlinespace

    Decoder layers & $\{2, 3, 4, \ldots, 12 \}$ & \\ \addlinespace

    Decoder FFN dim. & $\{ 512, 1024, 2048, 4096 \}$ & The dimension of the intermediate layer for the feedforward network in the decoder. \\ \addlinespace

    Decoder attention heads  &  $\{ 4, 8, 16, 32 \}$  & Number of attention heads in the decoder. \\ \addlinespace

    Dim. model   & $\{128, 256, 512, 1024 \}$ &  Excludes the layers listed above for which the dimension is set separately. \\ \addlinespace

    Dropout     & $[0.0, 0.6]$ &  The dropout probability in the embeddings, pooler, and encoder   for the fully connected layers. \\ \addlinespace
    \bottomrule
    \end{tabularx}
    \label{table:hypopt}
\end{table}

In order to ensure that the hyperparameters used for the model are suited to each task, we run hyperparameter optimization using Ray Tune \citep{Moritz2018-ns}.
For the document- and author-based splits, the tuning is run on the \emph{on reactions} split; for the time-based split, it is run on the 1996 cutoff split; and for the reaction-type splits, it is run on the \emph{base} split (see \S\ref{sect:namerxnSplitsMethods}).
The hyperparameters we tune, along with their given ranges, are shown in Table~\ref{table:hypopt};
when tuning, we use an ASHAScheduler \citep{Li2020-tu}, optimize for the loss on the validation set, and consider 100 potential different trials.
We stick to using models that are able to fit on a single GPU (we predominantly use NVIDIA RTX A5000 and NVIDIA GeForce RTX 3090 GPUs with approximately 24GB of memory).
Therefore, any hyperparameter combinations that cause the model to run out of GPU memory during training are discarded at the end.

In general, we use early stopping when training our final models for each experiment using our optimized hyperparameters.
Early stopping is done using the loss on our validation set; this does not always perfectly correlate with accuracy, but it is fast to compute due to the fact that it can be done on each token in parallel using teacher forcing. 
When evaluating our models, we use beam search (with a width of 5) and compare the predicted SMILES to the ground truth SMILES after canonicalization. 

We wish to stress that it is likely that the models we train do not obtain state-of-the-art accuracy on the tasks we consider.
Various avenues can be explored to improve performance further, such as considering larger multi-GPU models, performing more extensive hyperparameter tuning, training for different amounts of time, and using SMILES augmentation or Polyak averaging \citep[\S8.7.3]{Goodfellow2016-sv} (both of which have previously been shown to help similar models \citep{Schwaller2019-ra}).\footnote{Having said that, we ensured that our method was able to obtain comparable results with the Molecular Transformer on the augmented USPTO-MIT dataset \citep[Table 3]{Schwaller2019-ra}; here, we found our model obtained a top-1 accuracy of 87.1\% and a top-5 accuracy of 94.3\%  when trained for 500k iterations.}
However, obtaining the absolute best performing model is not the aim of our work, and during experiments with slightly different models, datasets, and even training regimes, we found that the qualitative trends we observed remained fairly consistent even if exact accuracy values differed.

\subsection{Creating \fig{fig:violinReactantVersusReaction}}
\label{sect:violingPlot}

\fig{fig:violinReactantVersusReaction} was created by calculating (for each test reaction) the average cosine distance (in reactant and reaction fingerprint space) to the nearest 5 neighbor reactions in the corresponding split's training set. To create this plot we used radius 2 Morgan fingerprints with 2048 bits. Reagents (i.e., molecules that did not contribute atoms to the product) were removed from the reaction when computing fingerprints (otherwise the fingerprints tended to represent the common solvents, catalysts, etc used rather than the more unique characteristics of the reactions). The reaction fingerprints were calculated by computing the product fingerprints and then subtracting from these the reactant fingerprints:
\begin{equation*}
    \textsf{Reaction Fingerprint} = \texttt{fingerprint}\big(\textsf{products} \big) - \texttt{fingerprint}\big( \textsf{reactants} \big).
\end{equation*}

Note that the different groups of splits in this plot have slightly different setups, for instance, the time splits use far smaller training set sizes than the NameRxn splits. Aside from the \emph{on reactions} split, all other splits use distinct documents for the train/test set.
``NameRxn ID'' is calculated with the in-distribution (ID) test set for the Grignard Ester split, but similar results are seen when using the ID test sets for the other NameRxn splits. Again note that this split is slightly different to the \emph{on documents} split as reactions with NameRxn class \nr{0.0} (i.e., uncategorized) are removed (see \S\ref{sect:namerxnSplitsMethods} for further details).

\clearpage
\section{Further experimental results}

This appendix contains further experimental results, complimenting those presented in the main text. 
Section~\ref{sect:tables} provides tables listing the main numerical results, while Section~\ref{sect:moreTimeSplits} provides further plots for the time-based splits.

\subsection{Tables}
\label{sect:tables}
\FloatBarrier

The tables provided here extend the main accuracy results presented in the main text.
In particular, Table~\ref{tab:docAuth} provides top-1 through top-5 accuracies for the document- and author-based splits (see \fig{fig:docAuth} and \S \ref{sect:docAuth}),
Table~\ref{tab:bh} provides the same for the Buchwald–Hartwig test set (see \fig{fig:bhOverTime} and \S \ref{sect:buchwaldHartwigAnalysis}), and Table~\ref{tab:namerxn} contains the accuracies for the NameRxn splits (see \fig{fig:namerxnSplits} and \S \ref{sect:nameRxn}).
\vspace{1cm}

\begin{table}[thp]  
\centering
\caption{Accuracies on the document- and author-based splits introduced in \S \ref{sect:docAuth}.
}
\label{tab:docAuth}
\begin{tabular}{llllll}
\toprule
\textbf{Split} & \textbf{top-1} & \textbf{top-2} & \textbf{top-3} & \textbf{top-4} & \textbf{top-5} \\
\midrule
on reactions & 0.65 & 0.72 & 0.74 & 0.76 & 0.77 \\
on documents & 0.58 & 0.65 & 0.68 & 0.69 & 0.70 \\
on authors & 0.55 & 0.62 & 0.64 & 0.66 & 0.67 \\
\bottomrule
\end{tabular}

\end{table}
\vspace{0.5cm}

\begin{table}[thp]  
\centering
\caption{Accuracies on the Buchwald–Hartwig test set for the models trained on different time splits (see \S\ref{sect:buchwaldHartwigAnalysis}). 
Note that the training set sizes for the time-based split models are smaller than those used for the other splits (as we control for the training set size across the different time cutoffs).
}
\label{tab:bh}
\begin{tabular}{rlllllr}
\toprule
\textbf{Split year} & \textbf{top-1} & \textbf{top-2} & \textbf{top-3} & \textbf{top-4} & \textbf{top-5} & \textbf{Num. of BH reactions
 in split's training set} \\
\midrule
1996 & 0.05 & 0.06 & 0.07 & 0.08 & 0.08 & 1 \\
1998 & 0.08 & 0.11 & 0.13 & 0.14 & 0.14 & 5 \\
2000 & 0.15 & 0.19 & 0.20 & 0.21 & 0.22 & 26 \\
2002 & 0.15 & 0.19 & 0.20 & 0.21 & 0.22 & 53 \\
2004 & 0.24 & 0.29 & 0.30 & 0.32 & 0.32 & 126 \\
2006 & 0.28 & 0.32 & 0.34 & 0.35 & 0.36 & 195 \\
2008 & 0.32 & 0.37 & 0.39 & 0.41 & 0.41 & 306 \\
2010 & 0.39 & 0.44 & 0.46 & 0.48 & 0.48 & 520 \\
2012 & 0.40 & 0.45 & 0.47 & 0.48 & 0.49 & 714 \\
2014 & 0.49 & 0.55 & 0.57 & 0.58 & 0.59 & 1071 \\
2016 & 0.55 & 0.60 & 0.62 & 0.63 & 0.64 & 1251 \\
2018 & 0.64 & 0.70 & 0.72 & 0.73 & 0.74 & 1684 \\
2020 & 0.62 & 0.68 & 0.69 & 0.70 & 0.71 & 1750 \\
\bottomrule
\end{tabular}

\end{table}

\begin{table}[thp]  
\centering
\caption{Accuracies on the reaction type splits introduced in \S \ref{sect:nameRxn}. 
Note that the rows marked ``w/1k OOD'' are for the baseline models, where 1000 OOD reactions are added to the original training set to give a sense of the intrinsic difficulty associated with that reaction class.
}
\label{tab:namerxn}
\begin{tabular}{llllll}
\toprule
\textbf{Split} & \textbf{top-1} & \textbf{top-2} & \textbf{top-3} & \textbf{top-4} & \textbf{top-5} \\
\midrule
Grignard Ester & 0.10 & 0.15 & 0.20 & 0.25 & 0.28 \\
Grignard Ester w/1k OOD & 0.85 & 0.89 & 0.91 & 0.92 & 0.93 \\
Chloro Suzuki & 0.74 & 0.82 & 0.85 & 0.87 & 0.88 \\
Chloro Suzuki w/1k OOD & 0.81 & 0.88 & 0.90 & 0.91 & 0.92 \\
Heck & 0.07 & 0.19 & 0.29 & 0.36 & 0.40 \\
Heck w/1k OOD & 0.63 & 0.81 & 0.86 & 0.88 & 0.89 \\
All Suzuki & 0.51 & 0.62 & 0.67 & 0.70 & 0.72 \\
All Suzuki w/1k OOD & 0.76 & 0.83 & 0.86 & 0.87 & 0.89 \\
Triflyloxy Suzuki & 0.83 & 0.90 & 0.92 & 0.93 & 0.94 \\
Triflyloxy Suzuki w/1k OOD & 0.85 & 0.91 & 0.93 & 0.93 & 0.94 \\
\bottomrule
\end{tabular}

\end{table}
\FloatBarrier

\subsection{Additional time-based split results}
\label{sect:moreTimeSplits}

\begin{figure}[ht]
    \centering
    \includegraphics[width=0.8\linewidth]{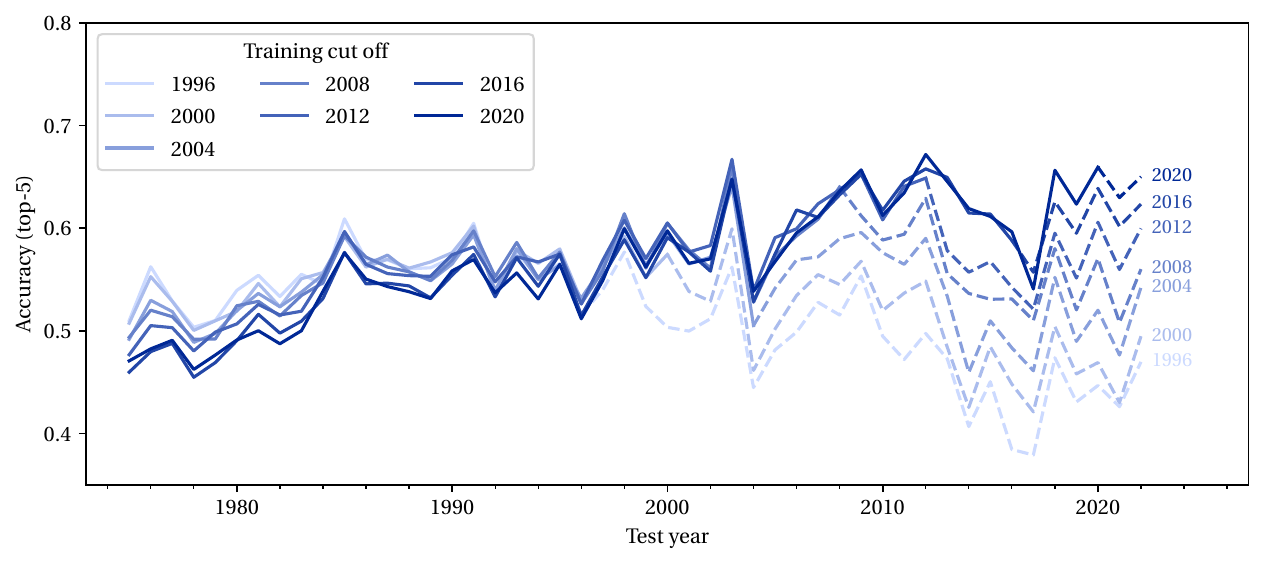}
    \caption{
    	Top-5 accuracy for reaction predictors trained up to different timepoints (different colors) when evaluated on held-out test sets for each year (x-axis). 
	For instance the line in the lightest shade, marked ``1996'', reports the top-5 accuracy for a reaction predictor trained on reactions that were reported up to and including 1996.
	The dashed line indicates model performance when the model is ``extrapolating'', in this context meaning that the test set year is beyond those associated with the reactions seen in the model's  training set.
	Similar to \fig{fig:timeSplits} in the main paper, we control for training set size (so each model sees the same number of reactions in training).
	}
    \label{fig:timeSplitstop5}
\end{figure}

This section contains more results for the time-based splits.
\fig{fig:timeSplitstop5} is the complement of \fig{fig:timeSplits} in the main text, showing the equivalent top-5 accuracy (the figure in the main text shows the top-1 accuracy) for the models trained on different time splits, when we control for training set size.

\begin{figure}[htp]
    \centering
    \includegraphics[width=0.8\linewidth]{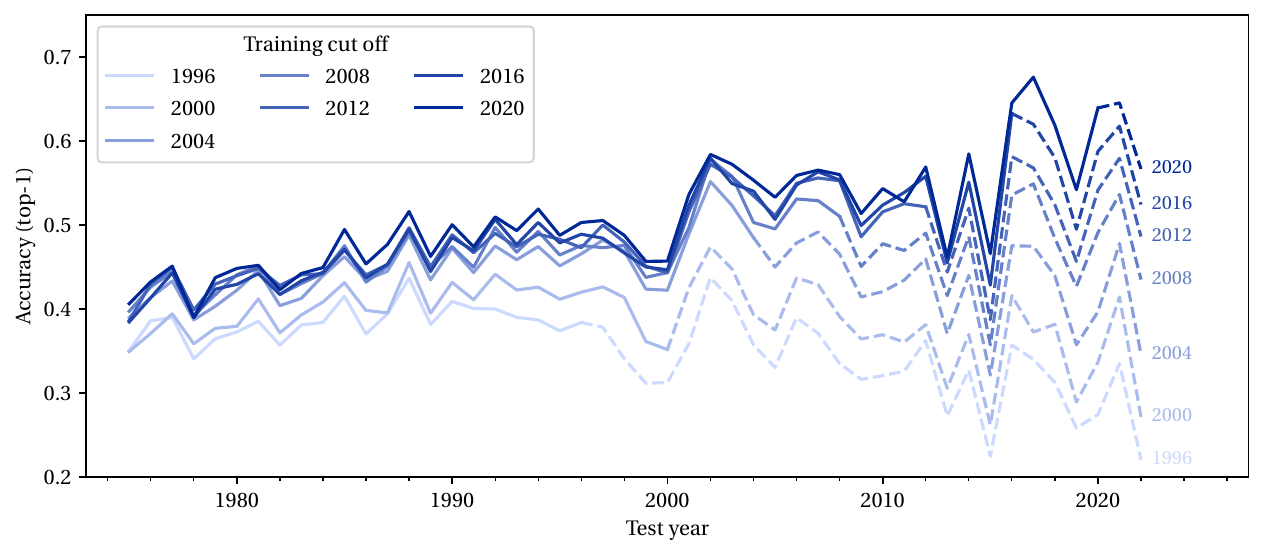}
    \caption{
    	Top-1 accuracy for reaction predictors trained up to different timepoints (different colors) when evaluated on held-out test sets for each year (x-axis). 
	For instance, the line in the lightest shade, marked ``1996'', reports the top-1 accuracy for a reaction predictor trained on reactions that were reported up to 1996 (inclusive).
	The dashed line indicates model performance when the model is ``extrapolating''—meaning that the test set year is beyond those associated with the reactions seen in the model's  training set.
	Note unlike \fig{fig:timeSplits} in the main paper, \textbf{we do not control for training set size} for these models (so each model sees a different number of reactions in training).
	}
    \label{fig:timeSplitsNoFix1}
\end{figure}

\begin{figure}[htp]
    \centering
    \includegraphics[width=0.8\linewidth]{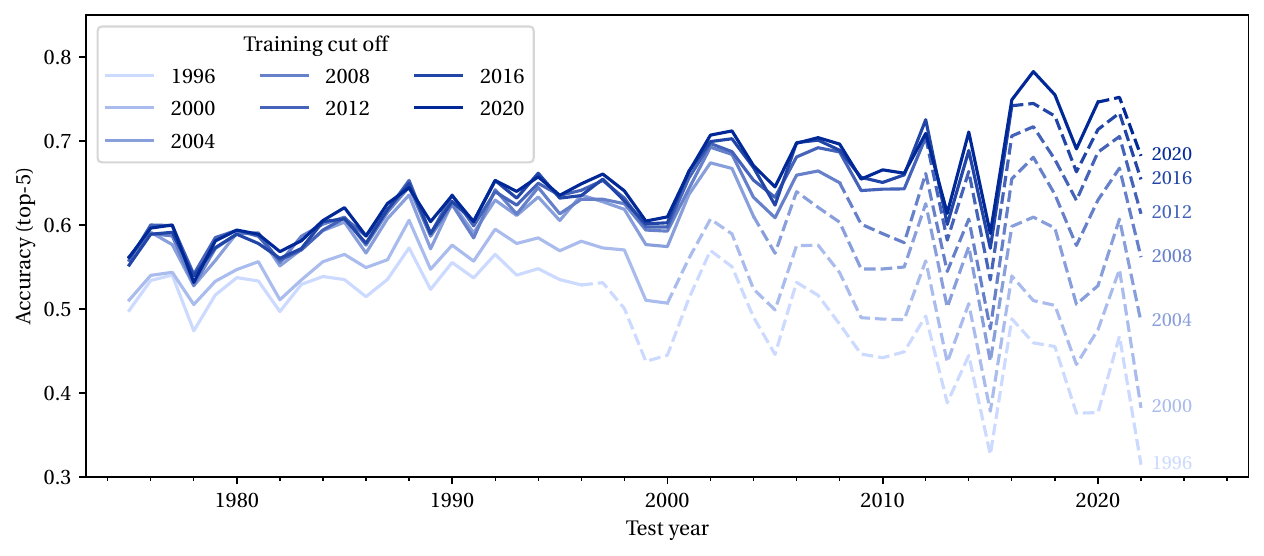}
    \caption{
    	Top-5 accuracy for reaction predictors trained up to different timepoints (different colors) when evaluated on held-out test sets for each year (x-axis). 
	For instance, the line in the lightest shade, marked ``1996'', reports the top-5 accuracy for a reaction predictor trained on reactions that were reported up to 1996 (inclusive).
	The dashed line indicates model performance when the model is ``extrapolating''—meaning that the test set year is beyond those associated with the reactions seen in the model's  training set.
	Note unlike \fig{fig:timeSplits} in the main paper, \textbf{we do not control for training set size} for these models (so each model sees a different number of reactions in training).
	}
    \label{fig:timeSplitsNoFix5}
\end{figure}

\fig{fig:timeSplitsNoFix1} and \fig{fig:timeSplitsNoFix5} show the top-1 and top-5 accuracies respectively for an experiment where we no longer control for training set size when forming the splits. 
Specifically, for each time cutoff, we use all of the available reactions (after removing  those used to form the held-out test sets) to train each model.
This means that models trained on sets associated with later time cutoffs will have seen far more reactions than models trained on the earlier ones.
While this approach may better reflect how these models  are updated in the real world, it entangles the effects of different dataset sizes with changing data distributions.
\vspace{0.5cm}

\end{appendices}

\end{document}